\documentclass[a4paper,conference]{IEEEtran}
\usepackage{booktabs}
\usepackage{dcolumn}
\usepackage{graphicx}

\usepackage{cite}

\ifCLASSINFOpdf
\else
\fi
\usepackage{amsmath}

\begin{document}
%
\title{Fast Motion Deblurring for Feature Detection and Matching Using Inertial Measurements}

\author{\IEEEauthorblockN{Janne Mustaniemi}
\IEEEauthorblockA{Center for Machine Vision\\
and Signal Analysis\\ 
University of Oulu, Finland\\
Email: janne.mustaniemi@oulu.fi}
\and
\IEEEauthorblockN{Juho Kannala\\ and Simo S\"arkk\"a}
\IEEEauthorblockA{Aalto University\\
Finland}
\and
\IEEEauthorblockN{Jiri Matas}
\IEEEauthorblockA{Centre for Machine Perception\\ 
Department of Cybernetics\\
Czech Technical University\\
Prague, Czech Republic}
\and
\IEEEauthorblockN{Janne Heikkil\"a}
\IEEEauthorblockA{Center for Machine Vision\\
and Signal Analysis\\ 
University of Oulu, Finland}}


%


\maketitle

\begin{abstract}
Many computer vision and image processing applications rely on local features. It is well-known that motion blur decreases the performance of traditional feature detectors and descriptors. We propose an inertial-based deblurring method for improving the robustness of existing feature detectors and descriptors against the motion blur. Unlike most deblurring algorithms, the method can handle spatially-variant blur and rolling shutter distortion. Furthermore, it is capable of running in real-time contrary to state-of-the-art algorithms. The limitations of inertial-based blur estimation are taken into account by validating the blur estimates using image data. The evaluation shows that when the method is used with traditional feature detector and descriptor, it increases the number of detected keypoints, provides higher repeatability and improves the localization accuracy. We also demonstrate that such features will lead to more accurate and complete reconstructions when used in the application of 3D visual reconstruction.
\end{abstract}


%
\IEEEpeerreviewmaketitle

\section{Introduction}
Feature detection and description form the basis of many computer vision and image processing applications. During past decades, various detectors and descriptors have been proposed, which have proven to be relatively stable against illumination changes, geometric transformations and image noise. Considerably less attention has been given to the problem of finding local correspondences in the presence of motion blur. Generally, the motion blur reduces the number of detected features, affects negatively the localization accuracy and makes the feature matching more difficult. The issue is most apparent in applications that involve a moving camera such as visual odometry, simultaneous localization and mapping (SLAM) and augmented reality (AR).

Image deblurring is one possible approach to address the problem of motion blur. Blind deconvolution is a process of recovering the sharp image and the point-spread-function (PSF) of the blur given an image. This is an ill-posed problem since the blurred image only provides a partial constraint on the solution. Blind deconvolution algorithms have significantly improved over the years, but they still fail frequently especially for larger blur sizes. What further complicates the problem is that motion blur often varies spatially across the image. For computational simplicity, existing algorithms typically assume spatially-invariant blur.

Mobile devices are commonly equipped with an inertial measurement unit (IMU). It provides additional information about the motion of the camera, which can be used to estimate the blur. If the PSF is known in advance, the problem of deblurring is known as non-blind deconvolution. Prior work, such as \cite{joshi2010image,vsindelavr2013image,hee2014gyro,hu2016image,zhang2016combining} has successfully used accelerometers and gyroscopes to improve deblurring. These methods can handle spatially-variant blur unlike most blind deconvolution methods. One major drawback with state-of-the-art deblurring algorithms is that they are computationally way too expensive to be used in real time applications such as SLAM.

In this paper, we propose an inertial-based deblurring method for improving the robustness against motion blur of existing feature detectors and descriptors. An implementation on GPU is capable of running in real time, unlike the state-of-the-art deblurring algorithms. It can handle spatially-variant blur and rolling shutter distortion. To account for the limitations of inertial-based blur estimation, the blur estimates are validated using image data. The evaluation shows that when the method is used together with traditional feature detector and descriptor, it increases the number of detected keypoints, provides higher repeatability and improves the localization accuracy of the detector. We also demonstrate applying our method for 3D visual reconstruction. The results show that it improves the accuracy of reconstructed 3D points and pose estimates; the number of reconstructed points also increases. Images that were originally too blurred to be registered are successfully matched.

\section{Related work}
Gauglitz et al. \cite{gauglitz2011evaluation} evaluated several interest point detectors and feature descriptors for visual tracking. The effect of motion blur was also studied. Fast Hessian \cite{bay2008speeded} and Difference of Gaussian (DoG) \cite{lowe1999object} detectors were found to be the most tolerant against motion blur as they provided the highest repeatability. Nevertheless, the performance of all detectors clearly degraded as motion blur increased. Neither of the detectors worked particularly well with strong motion blur. The evaluation of descriptors showed a similar decrease in performance, especially when matching blurred and non-blurred images. In such case, the SIFT \cite{lowe2004distinctive} outperformed the other descriptors.

To address the issue with motion blur, Pretto et al. \cite{pretto2007reliable,pretto2009visual} estimate the blur directly from the image. In \cite{pretto2007reliable}, the recovered PSF, which is assumed to be spatially-invariant and linear, is used to deblur the image before feature detection and description. The method in \cite{pretto2009visual} relaxes the assumption of spatially-invariant blur. Instead of deblurring, the estimated PSFs are used to build an adapted Gaussian scale-space pyramid. The idea is to blur less on the direction of the motion rather than applying an isotropic Gaussian blur to all levels in the pyramid.

A feature descriptor robust to different types of image blur was proposed in \cite{lee2016blur}. The descriptor is build by considering integral projections along different angular directions. According to the paper, the method outperforms the traditional descriptors. However, it still uses a feature detector that is not robust to motion blur. We argue in favor of deblurring, which allows us to use the existing feature detectors and descriptors that are known to perform well under various conditions.

In the context of visual SLAM,  motion blur is handled in  \cite{lee2011simultaneous,jin2005visual,russo2013blurring}. Lee et al. \cite{lee2011simultaneous} use the information from the SLAM system to estimate the motion blur. The estimates are then used for deblurring, although not all frames are deblurred. The methods \cite{jin2005visual,russo2013blurring} are based on a different idea. Instead of deblurring, the image patches are blurred to make the patches look similar. This may help in feature matching, but the number of detected features is usually greatly reduced in the presence of motion blur which is a problem especially when the number of distinctive features is limited. It is also worth mentioning that unlike  previous methods \cite{lee2011simultaneous,jin2005visual,russo2013blurring}, our approach, which operates on single images, is not limited to SLAM.

Visual-inertial SLAM systems such as \cite{leutenegger2015keyframe} can provide more accurate and reliable odometry information than purely vision-based approaches. The IMU can partially solve the issue of motion blur since it gives relatively accurate short-time estimates of the camera motion. However, the mapping does not take into account the motion blur, which can reduce the number of reconstructed 3D points and lower their accuracy. Relying on the IMU for longer periods of time will also cause drift.

\section{Blur estimation}
In this section, we describe how to estimate the motion blur from inertial measurements. Similar to many existing works, we only use gyroscope readings to recover the rotation of the camera during the exposure. It has been shown that rotation is typically the main cause of motion blur \cite{vsindelavr2013image,hee2014gyro}. Recovering the translation from the accelerometer readings is more difficult since it requires knowledge about the initial velocity of the camera. 

\subsection{Inertial Measurements}
Gyroscope measures the angular rate $\mathbf{\omega}(t)$ of the device in the sensor coordinate frame. The orientation of the device can be determined by integrating the angular velocities. It is well known that integration of noisy measurements can cause orientation to drift. In our case, we used a relatively short integration period to mitigate this problem (less than 33 ms). In the following sections, the camera orientations during the exposure are represented by rotation matrices $\mathbf{R}(t)$.

\subsection{Linear Motion Blur}
Due to short integration period, we can assume the motion blur is linear and homogeneous. This type of blur can be defined by angle $\theta$ and extent $r$. It is not necessary to consider the absolute motion of the camera when estimating the blur. Instead, we model the relative motion using a planar homography. If the camera is moving during the exposure, the 3D point is projected to multiple points, causing motion blur. Let $\mathbf{x}=(x,y,1)^\top$ be the projection of the 3D point at the beginning of the exposure in homogeneous coordinates. The homography

\begin{equation}
\mathbf{H}(t) = \mathbf{K} [ \mathbf{R}(t) + \frac{\mathbf{t}(t) \mathbf{n}^{\top}}{d} ] \mathbf{K}^{-1}
\end{equation}

maps the point to later time $t$, corresponding to the end of the exposure. The translation $\mathbf{t}$ is assumed to be zero so we can ignore the depth $d$ and the unit vector $\mathbf{n}$ that is orthogonal to the image plane. The mapping of points simplifies to 

\begin{equation}
\mathbf{x}' = \mathbf{K} \mathbf{R}(t) \mathbf{K}^{-1} \mathbf{x}.
\label{eq:homography1}
\end{equation}

The line segment connecting the points $\mathbf{x}$ and $\mathbf{x}'$ represents the linear motion blur. The angle $\theta$ and the extent $r$ of the blur can be obtained by

\begin{equation}
r = \sqrt{(x'-x)^2 + (y'-y)^2},
\end{equation}
\begin{equation}
\theta = \text{atan2} \bigg( \frac{y'-y}{x'-x} \bigg).
\end{equation}

It can be noted that we only have a discrete set of rotation matrices $\mathbf{R}(t)$. The intermediate rotations can be computed using the spherical linear interpolation (SLERP) \cite{shoemake1985animating}.

\subsection{The Rolling Shutter Effect}
When using a rolling-shutter camera, the rows in the image are captured at different time instances. This has to be taken into account when computing the motion blur since the mapping of the point $\mathbf{x}$ in equation (\ref{eq:homography1}) depends on its y-coordinate. Let $\mathbf{R}(t_1)$ and $\mathbf{R}(t_2)$ represent the camera orientations at the beginning and at the end of the exposure, respectively. We modify the equation (\ref{eq:homography1})

\begin{equation}
\mathbf{x}' = \mathbf{K}\mathbf{R}(t_2)\mathbf{R}^{\top}(t_1)\mathbf{K}^{-1}\mathbf{x},
\label{eq:homography2}
\end{equation}

where $t_1$ and $t_2$ are computed as follows. Time difference between the exposure of the first and last row of the image is defined by the camera readout time $t_r$. Let $t_f$ be the frame timestamp, that is the start of the first row exposure. Then, the exposure of the $y$:th row starts at

\begin{equation}
t_1(y) = t_f + t_r \frac{y}{N},
\end{equation}

where $N$ is the number of rows in the image. Given the exposure time $t_e$, the end of the exposure is defined as $t_2 = t_1 + t_e$. The timestamp $t_f$, readout time $t_r$ and exposure time $t_e$ can be typically obtained via the API of the mobile device.

The rolling shutter not only affects the blur estimation but it also creates a geometrical distortion to the image. Unlike existing rectification methods such as \cite{forssen2010rectifying,karpenko2011digital}, we do not rectify the whole image, which can be computationally expensive. Instead, we only need to refine the feature locations using the formula (\ref{eq:homography1}).

\subsection{Blur Validation}
Previously, we used gyroscope readings to estimate the blur angle $\theta$ and extent $r$. Even though the motion blur is mainly caused by rotation, translation can still cause problems in certain situations. For example, when the camera orbits around the object, the image may appear sharp while the gyroscope still measures rotation. Consequently, the blur extent gets over-estimated. We want to recognize this situation and avoid deblurring an already sharp image. To achieve this, we utilize the image data, which also contains information about the blur.

To check whether we should trust the blur estimates from the gyroscope, we compute the gradient magnitudes along the motion direction. If the blur estimate is correct, the image (or image patch) should not have strong gradients along this direction. On the other hand, strong gradients indicate that blur estimate should not be trusted. To compute the gradients, we convolve the image with a Sobel filter that has been rotated by blur angle $\theta$. If the maximum gradient exceeds a threshold $\tau$, the safest approach is not to deblur the image. In such case, we also do not rectify the feature locations.

\section{Deblurring}
In this section, we perform spatially-variant deblurring using Wiener deconvolution filter. Deconvolution is performed in the spatial domain, which has advantages over the conventional frequency domain approach. Most importantly, it allows real time performance. To handle spatially-variant blur, the image is first divided into smaller blocks as shown in Figure \ref{fig:realworld}. Then, the blur parameters $\theta$ and $r$ are estimated for the center pixel of each block. The spatial deconvolution is performed separately for each block.

\subsection{Spatial Deconvolution}
\begin{figure*}
 \includegraphics[width=1\textwidth]{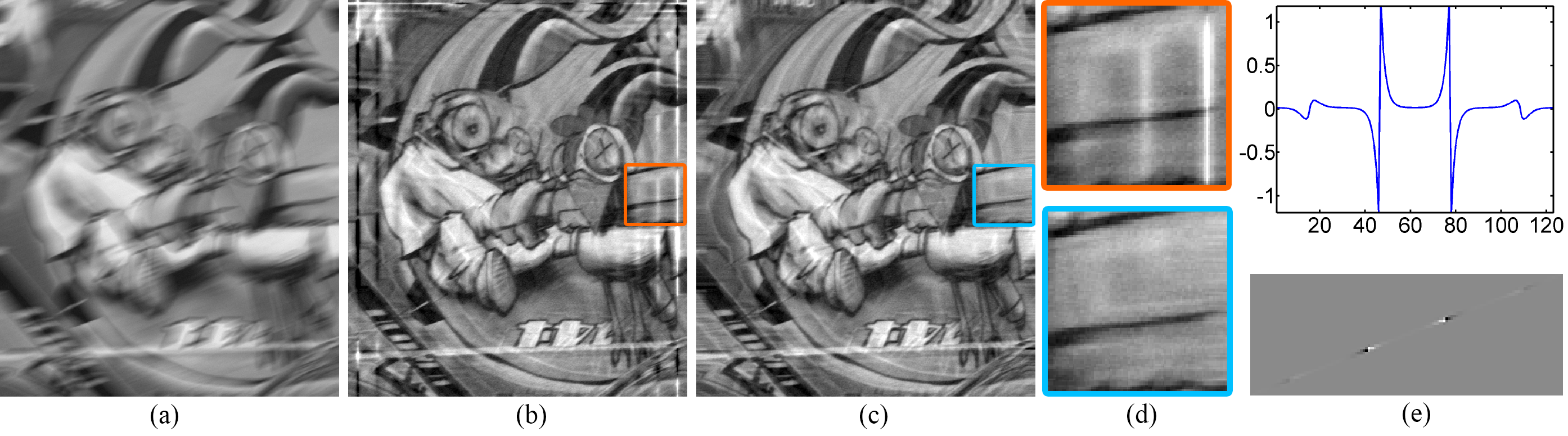}
\caption{Deconvolution in the frequency domain vs. spatial domain. (a) Image with synthetic motion blur ($\theta = 20$ degrees, $r = 31$ pixels), (b) Wiener deblurred image (frequency domain), (c) Wiener deblurred image (spatial domain), (d) Details showing the edge artifacts common to frequency domain approach, (e) Spatial deconvolution kernels in 1D and 2D.}
\label{fig:deconvolution}
\end{figure*}

Wiener deconvolution is commonly performed in the frequency domain. An example of a deblurred image is shown in Figure \ref{fig:deconvolution}(b). Notice the ringing artifacts near the edges of the image. These artifacts could be suppressed by applying a prepocessing step known as edge tapering. However, in the spatial domain approach this operation is not needed.

Let $h(x)$ represent a 1D motion blur kernel and $H(f) = \mathcal{F} \{ h(x) \}$ its Fourier transform. The Wiener deconvolution filter is defined as

\begin{equation}
W(f)=\frac{H^\star(f)}{|H(f)|^2 + \gamma},
\end{equation}

where $H^\star(f)$ is the complex conjugate of $H(f)$. Regularization term $\gamma$ represents the noise-to-signal ratio, which helps to suppress the high frequency components of the inverse filter.

The spatial counterpart of the filter can be computed by taking the inverse Fourier transform $w(x) = \mathcal{F}^{-1} \{ W(f) \}$. An example of a spatial deconvolution kernel is shown at the top of the Figure \ref{fig:deconvolution}(e). We can see that inverse kernel has a compact support due to regularization,  the length of the inverse kernel depends on the blur extent. In our experience, we get satisfactory deblurring results when the length of the inverse kernel is four times the blur extent $r$ (assuming $\gamma = 0.01$). This is achieved by padding both sides of $h(x)$ with zeros before taking the Fourier transform. The final 2D deconvolution kernel is obtained by rotating the 1D kernel by angle $\theta$. An example of a blurred image is shown in Figure \ref{fig:deconvolution}(c). Compared to frequency domain approach, there are no significant ringing artifacts near the edges of the image.

\subsection{Fast Implementation}
The spatial deconvolution kernels can be quite large, which is the main reason why spatial domain is usually not preferred. However, most of the kernel elements will be zeros when the blur is linear. See the gray pixels at the bottom of the Figure \ref{fig:deconvolution}(e). This leads to significant performance improvement since we only need to process nonzero elements. Furthermore, these kernels can be computed offline given a set of blur angles and extents (e.g. $\theta = 0,1,...,179$ and $r = 2,3,...,r_{max}$). The frequency domain approach requires computing the Fourier transforms at runtime, which quickly becomes time consuming, especially when the blur is spatially-variant.

THe spatial domain approach is also easy to parallelize, which allows a fast GPU implementation. In our experiments, we used the NVIDIA GeForce GTX 1080 GPU. It takes approximately 17 milliseconds to deblur a single grayscale image with resolution of 1920 x 1080 pixels when the average blur extent is 90 pixels. The running time could be further optimized, for example by utilizing the local memory resources of the GPU. 

\section{Experiments}
Algorithms were evaluated on both synthetically and naturally blurred images. All experiments were performed using the publicly available DoG + SIFT implementation \cite{siftcode}, although the proposed method can be used with any feature detector and descriptor. The proposed method is compared against the original implementation \cite{siftcode} and Pretto et al. \cite{pretto2009visual}, which we have reimplemented. Instead of estimating the blur parameter from the image as in \cite{pretto2009visual}, we use our inertial-based blur estimates. In the last section, we also demonstrate using our approach in the application of 3D visual reconstruction.

\subsection{Synthetic Blur}
In this experiment, we add synthetic motion blur and 30 dB Gaussian noise to the standard test images \cite{mikolajczyk2005performance}. These images contain viewpoint changes, scale changes, out-of-focus blur, JPEG compression and illumination changes. Since we are mainly interested in the effects of motion blur, we use the first image pair from each of the 8 image sets for which the above mentioned transformations are modest. The blur parameters are set to $\theta = 30$ degrees and $r = 27$ pixels for the first image and $\theta = 110$ degrees and $r = 47$ pixels for the second image.

\begin{figure*}
 \includegraphics[width=1.0\textwidth]{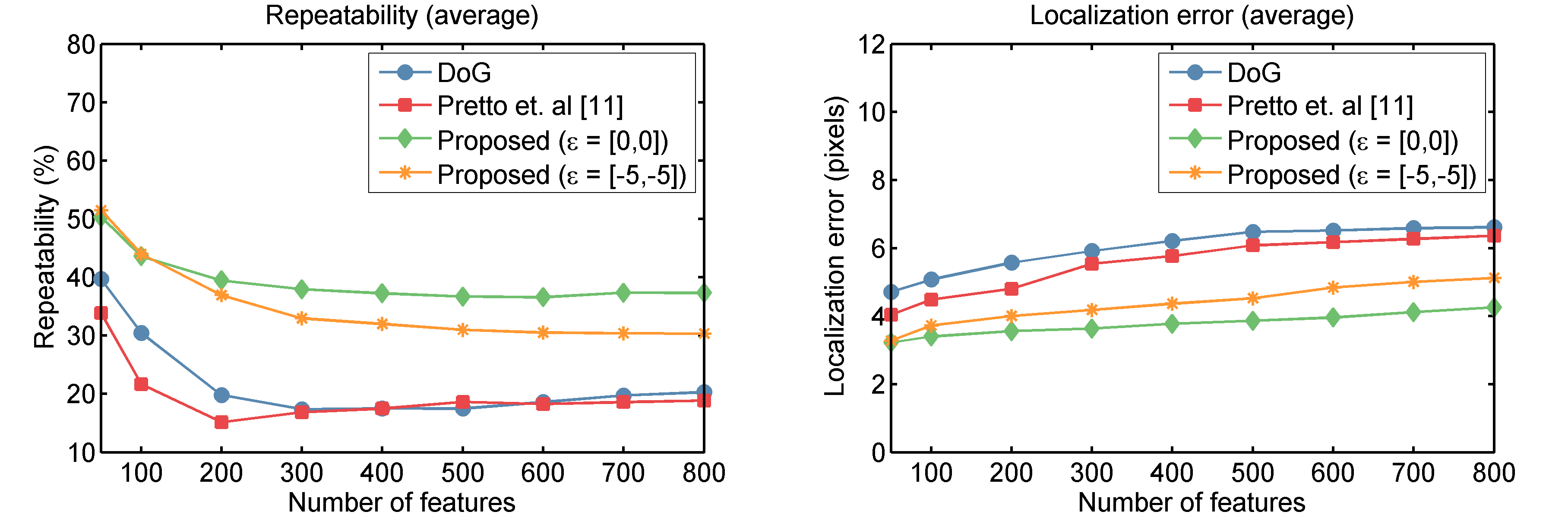}
\caption{Results on synthetically blurred images. Average repeatability and localization errors are computed for a fixed number of detections (x-axis). The localization error is the average distance between corresponding keypoints (passing the 40 \% overlap criteria) after reprojecting them to the reference image. The error of the blur estimate is denoted by $\epsilon$. For example, $\epsilon = [-5,-5]$ means that instead of using the ideal blur estimate for deblurring, the blur angle and extent have errors of -5 degrees and -5 pixels, respectively. In case of Pretto et al. \cite{pretto2009visual}, we use the ideal blur estimate.}
\label{fig:rep}
\end{figure*}

Repeatability and localization errors for different methods are shown in Figure \ref{fig:rep}. Results are averaged for all image pairs. We have fixed the number of detections by adjusting the respective threshold parameter. This eliminates the issue that repeatability criteria might favor those methods that return many keypoints such as ours. Although the method proposed by Pretto et al. \cite{pretto2009visual} is also capable of detecting more keypoints than the standard DoG, not all of them are distinctive. This problem was also recognized in the original paper. To remedy this issue, they used an addition step to discard less distinctive keypoints based on the entropy of the descriptors. We did not implement this step since the focus of this experiment was purely on keypoint detection. Overall, the proposed method outperforms the other methods in both repeatability and localization accuracy. The reason for better localization accuracy could be explained by the scale of the detections. Given a blurred image, the standard DoG is likely to detect more large scale keypoints. Such detections will pass the 40 $\%$ overlap criteria more easily even if their distance in the reference image is large.

\subsection{Real-world blur}
\begin{figure*}
 \includegraphics[width=1.0\textwidth]{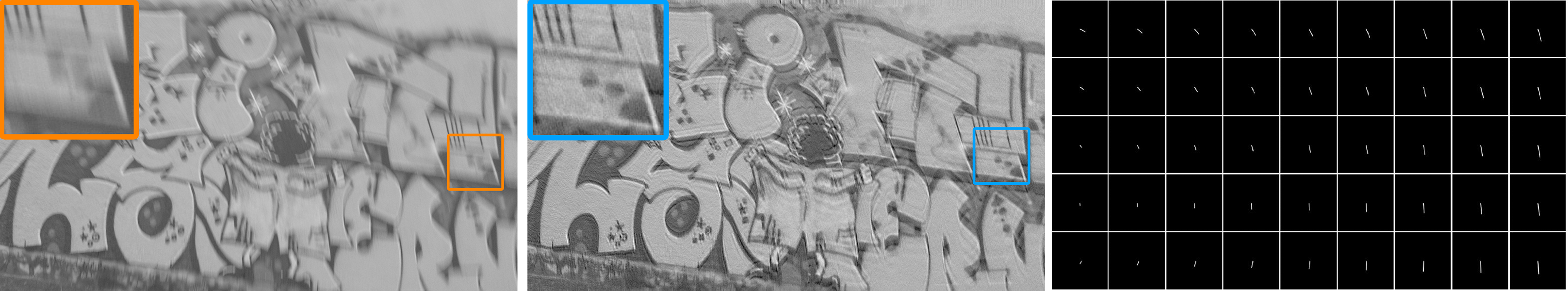}
\caption{A real-world image with spatially-variant motion blur (left). Deblurred image (center).  Blur kernels calculated from IMU motion estimates (right).}
\label{fig:realworld}
\vspace{3mm}
\end{figure*}

Naturally blurred images were captured with the NVIDIA Shield tablet while simultaneously recording gyroscope readings at 100 Hz. The dataset consists of two planar scenes, each containing 5 pairs of motion blurred images with resolution of 1080 x 1920 pixels. In many cases, the motion blur is spatially-variant as shown in Figure \ref{fig:realworld}.

To evaluate detector's performance, we need to know the homographies between the images, i.e. the ground truth mapping of image points in the first and second image. Usually, the estimation is done by carefully selecting corresponding image points from the images. However, selecting the points accurately is not an easy task when images contain motion blur. Instead of capturing a single motion blurred image, we capture a burst of three images at the rate of 30 fps while alternating short and long exposure time. For the first and last image of the sequence, we set the exposure time to 1 ms, which produces sharp but noisy images. The exposure time of the middle image was set to 30 ms causing motion blur.

Given the three consecutive images, we first detect a set of SURF keypoints \cite{bay2008speeded} from the sharp images and filter out the outliers with RANSAC. Due to short time difference between the captures, we assume that motion of the points is linear between the images. The corresponding keypoints in the middle image are obtained via linear interpolation. The process is repeated for the second viewpoint, after which the ground truth homography is estimated from the interpolated keypoints while utilizing RANSAC.

Results for the real-world experiments are shown in Table \ref{tab:realworld}. Here, the number of detections was fixed to 500. In almost all of the cases, the proposed approach achieves higher repeatability and lower localization error than other methods. Compared to the DoG, the performance is also more consistent. It is worth to mention that deblurring usually causes ringing artifacts, which can increase the number bad detections. However, this did not seem to be much of a problem on our datasets.

\begin{table}
\newcolumntype{.}{D{.}{.}{-1}}
\centering
\setlength{\tabcolsep}{9pt}
\caption{Results on the real-world data. Repeatability (rep.) and average localization error in pixels (err.) were computed using a fixed number of detections (500).}
\begin{tabular}{lllllll}
\toprule
\multicolumn{1}{c}{} &
\multicolumn{2}{c}{DoG \cite{siftcode}} &
\multicolumn{2}{c}{Pretto et al. \cite{pretto2009visual}} &
\multicolumn{2}{c}{Proposed} \\
\cmidrule(r){2-3}
\cmidrule(r){4-5}
\cmidrule(r){6-7}
\multicolumn{1}{l}{Graffiti 1} & 
\multicolumn{1}{l}{Rep.} &
\multicolumn{1}{l}{Err.} &
\multicolumn{1}{l}{Rep.} &
\multicolumn{1}{l}{Err.} &
\multicolumn{1}{l}{Rep.} &
\multicolumn{1}{l}{Err.} \\
\cmidrule(r){2-3}
\cmidrule(r){4-5}
\cmidrule(r){6-7}
\#1  & 18.4 &  7.1 &  42.9 &  3.6 &  \textbf{55.7} &  \textbf{3.4} \\
\#2  & 18.2 &  6.7 &  29.5 &  4.5 &  \textbf{47.5} &  \textbf{4.2} \\
\#3  & 18.3 &  6.7 &  32.0 &  6.3 &  \textbf{46.1} &  \textbf{4.0} \\
\#4  & 5.4 &  6.0 &  15.5 &  5.6 &  \textbf{39.0} &  \textbf{4.3} \\
\vspace{2mm}
\#5  & 21.1 &  6.5 &  37.5 &  3.8 &  \textbf{55.9} &  \textbf{3.6} \\
Graffiti 2  &   &    &    &    &    &    \\
\#1  & 18.0 &  7.6 &  36.8 &  \textbf{4.4} &  \textbf{45.4} &  4.8 \\
\#2  & \textbf{65.9} &  3.9 &  62.7 &  3.7 &  62.9 & \textbf{3.5} \\
\#3  & 55.8 &  6.2 &  28.3 &  4.8 &  \textbf{70.7} &  \textbf{3.8} \\
\#4  & 36.1 &  7.4 &  41.5 &  \textbf{3.7} &  \textbf{54.4} &  3.8 \\
\#5  & 10.7 &  6.8 &  28.5 &  3.9 &  \textbf{50.2} &  \textbf{3.7} \\
\bottomrule
\rule{0pt}{2ex}  
avg.  & 26.8 &  6.5 &  35.5 &  4.4 &  \textbf{52.8} &  \textbf{3.9} \\
\end{tabular}
\label{tab:realworld}
\end{table}

\subsection{Visual Reconstruction}
In this experiments, we use the proposed method for 3D visual reconstruction. We utilize the VisualSFM software \cite{wu2011visualsfm} with our inertial-aided features. For the evaluation, we captured a short video sequence consisting of 150 frames.

\begin{figure*}
 \includegraphics[width=1.0\textwidth]{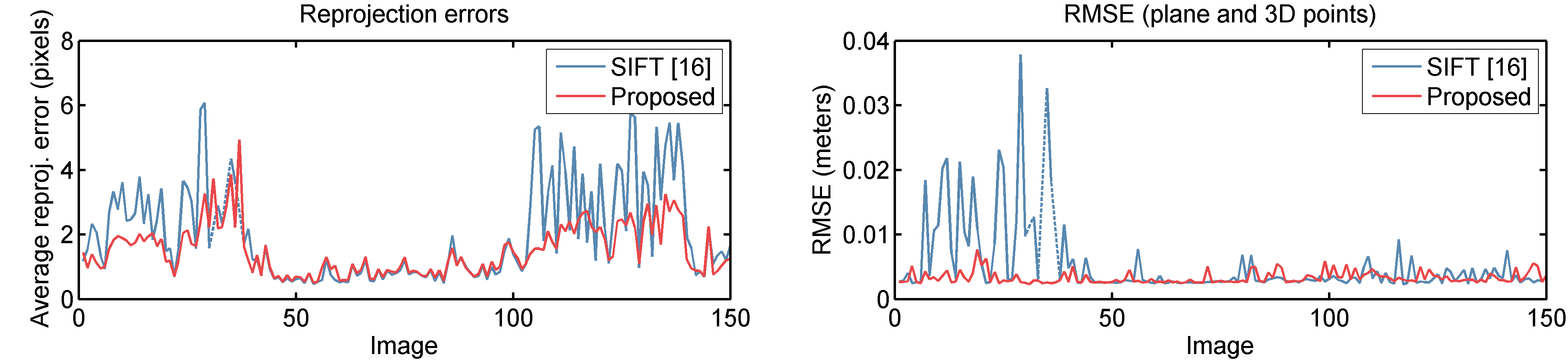}
\caption{Average reprojection errors and root-mean-square errors between the 3D points and the plane.}
\label{fig:reproj}
\end{figure*}

The captured scene is planar, which means all reconstructed 3D points should lie close to a plane. To check whether this is true, we first fit a plane to the points with RANSAC. Then, for the 3D points observed in each frame, we compute the root-mean-square error (RMSE) between the plane and the points. We also compute the average reprojection errors for each frame. It is assumed that all 3D points are visible at least in three images. Results are shown in Figure \ref{fig:reproj}. We can clearly see that our method outperforms the standard SIFT features. Frames at the beginning and end of the sequence are heavily blurred, which appear as spikes in the graphs. The standard SIFT is also not able to register all the images. There are three frames which were not registered due to strong motion blur. Furthermore, our method increases the total number of reconstructed points from 5888 points to 10620 points compared to the standard SIFT.

\begin{figure*}
 \includegraphics[width=1.0\textwidth]{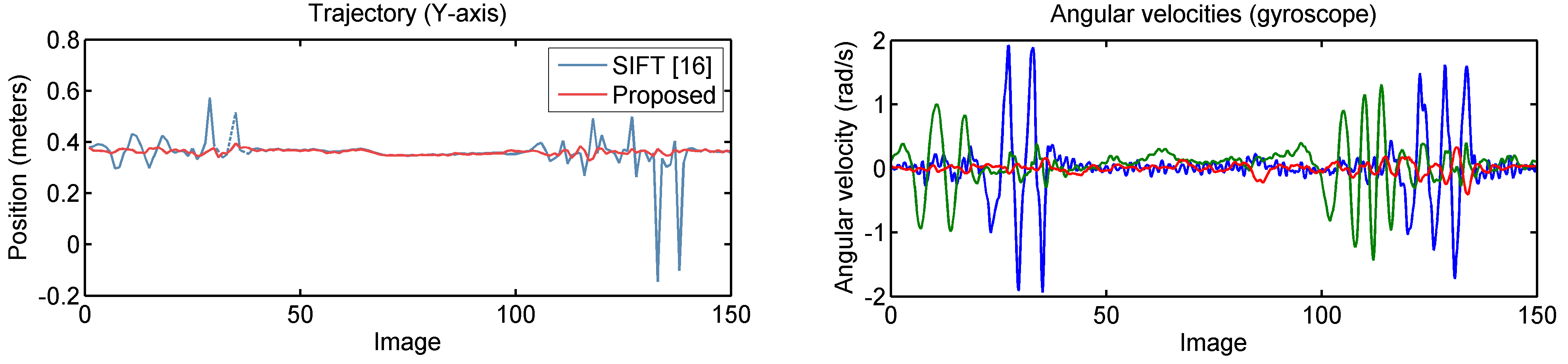}
\caption{Estimated camera height (``Y-axis'') and angular velocities measured by a gyroscope.}
\label{fig:trajectory}
\end{figure*}

The estimated camera trajectory is shown in Figure \ref{fig:trajectory}. Even though the ground truth trajectory is unknown, it is obvious that motion blur causes problems for the standard SIFT. There are unrealistic spikes in the trajectory, which appear in the presence of motion blur. It is also clear that these errors are highly correlated with angular velocities measured by the gyroscope.

\section{Conclusion}
We proposed an inertial-based deblurring method for feature detection and matching in the presence of motion blur. It is the first method capable of processing every frame of a high-definition video in real-time, which makes it suitable for applications such as SLAM. The method can also handle spatially-variant blur and rolling shutter distortion. The limitations of inertial-based blur estimation are taken into account by validating the blur estimates using image data. The evaluation shows that when the method is used with a traditional feature detector and descriptor, it increases the number of detected keypoints, provides higher repeatability and improves the localization accuracy of the detector. When applied to 3D visual reconstruction, the method improved the reconstruction quality. The paper focused on feature detection and matching, however, real-time deblurring has other applications as well.


\newpage
\section*{Acknowledgment}
The work has been financially supported by the FiDiPro programme of Business Finland and J. Matas was supported by Czech Science Foundation Project GACR P103/12/G084.



\bibliographystyle{IEEEtran}
%




\end{document}